\documentclass[lettersize,journal]{IEEEtran}
\usepackage{amsmath,amsfonts}
\usepackage{algorithmic}
\usepackage{algorithm}
\usepackage{array}
\usepackage[caption=false,font=normalsize,labelfont=sf,textfont=sf]{subfig}
\usepackage{textcomp}
\usepackage{stfloats}
\usepackage{url}
\usepackage{verbatim}
\usepackage{graphicx}
\usepackage{cite}

\usepackage{booktabs}
\usepackage{array}

\usepackage{svg}
\usepackage{multirow}
\usepackage{makecell}

\usepackage{siunitx}

\usepackage{hyperref}

\usepackage{orcidlink}

\usepackage[acronym]{glossaries}

\newacronym{ad}{AD}{Autonomous Driving}
\newacronym{il}{IL}{Imitation Learning}
\newacronym{rl}{RL}{Reinforcement Learning}
\newacronym{rlfp}{RLfP}{Reinforcement Learning from Pixels}
\newacronym{td}{TD}{Temporal Differences}
\newacronym{alix}{A-LIX}{Adaptive Local Signal Mixing}
\newacronym{ppo}{PPO}{Proximal Policy Optimization}
\newacronym{a3c}{A3C}{Asynchronous Advantage Actor Critic}
\newacronym{sac}{SAC}{Soft Actor-Critic}
\newacronym{ddpg}{DDPG}{Deep Deterministic Policy Gradient}
\newacronym{ema}{EMA}{Exponential Moving Average} 

\newcommand{\etal}{\textit{et al}., }
\newcommand{\ie}{\textit{i}.\textit{e}., }

\begin{document}

\title{RLAD: Reinforcement Learning from Pixels for Autonomous Driving in Urban Environments}

\author{Daniel Coelho \orcidlink{0000-0002-5743-7663} , Miguel Oliveira \orcidlink{0000-0002-9288-5058}, and Vítor Santos \orcidlink{0000-0003-1283-7388}%
\thanks{Daniel Coelho is with the Department of Mechanical Engineering, University of Aveiro, 3810-193 Aveiro, Portugal, and with the Intelligent System Associate Laboratory (LASI), Institute of Electronics and Informatics Engineering of Aveiro (IEETA), University of Aveiro, 3810-193 Aveiro, Portugal (e-mail: danielsilveiracoelho@ua.pt).}
\thanks{Miguel Oliveira is with the Department of Mechanical Engineering, University of Aveiro, 3810-193 Aveiro, Portugal, and with the Intelligent System Associate Laboratory (LASI), Institute of Electronics and Informatics Engineering of Aveiro (IEETA), University of Aveiro, 3810-193 Aveiro, Portugal (e-mail:mriem@ua.pt)}
\thanks{Vítor Santos is with the Department of Mechanical Engineering, University of Aveiro, 3810-193 Aveiro, Portugal, and with the Intelligent System Associate Laboratory (LASI), Institute of Electronics and Informatics Engineering of Aveiro (IEETA), University of Aveiro, 3810-193 Aveiro, Portugal (e-mail:vitor@ua.pt)}
}



\maketitle

\begin{abstract}
Current approaches of \acrfull{rl} applied in urban \acrfull{ad} focus on 
decoupling the perception training from the driving policy training. The main reason is to avoid training a convolution encoder alongside a policy network, which is known to have issues related to sample efficiency, degenerated feature representations, and catastrophic self-overfitting. However, this paradigm can lead to representations of the environment that are not aligned with the downstream task, which may result in suboptimal performances. To address this limitation, this paper proposes RLAD, the first \acrfull{rlfp} method applied in the urban \acrshort{ad} domain. We propose several techniques to enhance the performance of an \acrshort{rlfp} algorithm in this domain, including: i) an image encoder that leverages both image augmentations and \acrfull{alix} layers; ii) WayConv1D, which is a waypoint encoder that harnesses the 2D geometrical information of the waypoints using 1D convolutions; and iii) an auxiliary loss to increase the significance of the traffic lights in the latent representation of the environment.
 Experimental results show that RLAD significantly outperforms all state-of-the-art \acrshort{rlfp} methods on the NoCrash benchmark. We also present an infraction analysis on the NoCrash-regular benchmark, which indicates that RLAD performs better than all other methods in terms of both collision rate and red light infractions.

\end{abstract}


\section{Introduction} \label{sec:introduction}


In recent years, \acrfull{ad} has experienced significant growth due to advancements in artificial intelligence and information sensing, which have received widespread attention in both academia and industry \cite{FERNANDES2021161,Coelho2022}. In general terms, \acrshort{ad} involves tasks that fall into two main categories: environment perception and driving policy \cite{Wu2021, Chitta2021NEATNA, HUANG2023101834}. First, the autonomous agent must derive a useful representation of the environment from sensor data, and then generate the appropriate control commands based on the driving policy in order to keep the vehicle on a safe route.


Urban driving is one of the most challenging environments for autonomous vehicles, mainly due to the unpredictability and diversity of agents present in the environment, as well as complex situations, such as pedestrians crossing lanes, traffic lights, intersections, among others \cite{Toromanoff2019EndtoEndMR, Coelho2022}. Given the multifaceted nature of urban driving, researchers have recognized the necessity of employing information fusion techniques to effectively perceive and comprehend data from various sources and of different natures \cite{HUANG2023101834, OLIVEIRA2015108}. Consequently, there has been a notable shift in focus from traditional modular pipelines to end-to-end approaches, such as \acrfull{il} and \acrfull{rl} \cite{Ahmed2022}.

\acrfull{il} learns a driving policy from a dataset of expert demonstrations using supervised learning techniques \cite{Agarwal2021}, in which the goal is to create an agent that behaves as similarly as possible to the expert. The major limitations of this method are that the driving policy is limited to the performance of the experts, and it is practically inconceivable to collect expert data covering all possible driving situations \cite{Coelho2022}. \acrshort{il} algorithms also suffer from a distribution mismatch in the training data, \ie the algorithm will never encounter failing situations, and therefore, will not react properly in those conditions \cite{Toromanoff2019EndtoEndMR}. 

Conversely, \acrfull{rl} learns a driving policy by interacting directly with an environment and collecting rewards that assess the suitability of an action taken in a given state \cite{Chekroun2021GRIGR}. Usually, the goal of the agent is to maximize the cumulative rewards. As in this case the agent is interacting directly with the environment, it does not suffer from a distribution mismatch and is also not limited to the performance of an expert. However, due to the extensive exploration of the environment during the training stage, \acrshort{rl} is known to have a poor sample efficiency, requiring an order of magnitude more data than \acrshort{il} to converge \cite{Chekroun2021GRIGR}. 

\acrfull{rlfp} is a type of \acrshort{rl} that directly maps the image data into actions. This requires simultaneously training a convolution encoder alongside a policy network, which is a challenging task due to the sample efficiency problem \cite{Kostrikov2020ImageAI}. Additionally, it is known that performing \acrfull{td} learning with a convolutional encoder leads to unstable training and premature convergence, which eventually results in degenerated feature representations \cite{Cetin2022StabilizingOD}. 

Existing \acrshort{rlfp} approaches have been applied on Atari games \cite{Mnih2013PlayingAW} and MuJoCo \cite{6386109} tasks, which present significantly fewer challenges in terms of environment perception when compared to \acrshort{ad}. For instance, in Atari and MuJoCo, practically any change in the observation space is task-relevant, whereas in \acrshort{ad} the observation space contains predominately task-irrelevant information, as is the case of clouds and architectural details \cite{Zhang2020LearningIR}. To bypass this problem, current \acrshort{rl} approaches applied in urban \acrshort{ad} focus on decoupling the perception training from the driving policy training \cite{Zhao2022CADREAC, Huang2021, Toromanoff2019EndtoEndMR, Chekroun2021GRIGR, Agarwal2021, Ahmed2022}. The idea is to train an encoder using supervised or unsupervised techniques to derive a latent representation from the sensor data, and then train an \acrshort{rl} algorithm that maps the latent representation into actions. This adds stability to the optimization by circumventing dueling training objectives. However, it leads to suboptimal policies because the encoder may not be aligned with the downstream task \cite{Yarats2019ImprovingSE}. Since the objective is to maximize the cumulative rewards, it is beneficial to use them to improve simultaneously the feature representation of the sensor data and the driving policy network \cite{Zhang2020LearningIR}.

This paper proposes \textbf{RLAD}, a \textbf{R}einforcement \textbf{L}earning from Pixels \textbf{A}utonomous \textbf{D}riving agent, capable of driving under complex urban environments. This is the first approach to carry out a successful simultaneous training of the encoder and policy network using \acrshort{rl} in the domain of vision-based urban \acrshort{ad}. We leverage the latest advancements in \acrshort{rlfp} that have been achieved by Meta AI\footnote{\url{https://ai.facebook.com/}} and propose techniques to integrate those advancements in the urban \acrshort{ad} domain. Overall, we summarize our main contributions as follows:

\begin{itemize}
    \item We propose RLAD, the first method that learns simultaneously the encoder and the driving policy network using \acrshort{rl} in the domain of vision-based urban AD. We also show that RLAD significantly outperforms all state-of-the-art \acrshort{rlfp} methods in this domain;
    \item We introduce an image encoder that leverages both image augmentations and \acrfull{alix} layers to minimize the catastrophic self-overfitting of the encoder;
    \item We propose WayConv1D, a waypoint encoder that leverages the 2D geometrical information of the waypoints using 1D convolutions with a 2$\times$2 kernel, which significantly improves the stability of the driving;
    \item We perform a comparative analysis of the state-of-the-art \acrshort{rlfp} in the domain of vision-based urban \acrshort{ad}, where we show that one of the main challenges is obeying traffic lights. To address this limitation, we  incorporate an auxiliary loss that specifically targets the traffic light information in the latent representation of the image, thereby enhancing its significance.

\end{itemize}

\section{Related Work}

\subsection{Reinforcement Learning for Autonomous Driving}


\acrshort{rl} has been used in \acrshort{ad} to overcome the limitations of IL, however, vision-based \acrshort{rl}, or more precisely \acrshort{rlfp}, comes with several drawbacks \cite{Chekroun2021GRIGR}. Camera images are of high dimensions, thus requiring larger \acrshort{rl} networks and optimizing dueling training objectives: the image encoder and the policy network \cite{Yarats2019ImprovingSE}. To overcome these limitations, the common approach is to disentangle the perception network from the policy network and perform a two-stage training \cite{Zhao2022CADREAC, Huang2021, Toromanoff2019EndtoEndMR, Chekroun2021GRIGR, Agarwal2021, Ahmed2022}. The first stage consists of encoding the sensor data in a latent representation by pretraining a visual encoder on visual tasks, such as classification and segmentation \cite{Chekroun2021GRIGR}. Then, the latent representations are received by an \acrshort{rl} algorithm to train the driving policy network. Following this line, Toromanoff \etal \cite{Toromanoff2019EndtoEndMR} proposed a method called \textit{Implicit Affordances}. First, a visual encoder is trained using auxiliary tasks, such as traffic light state and distance, road type, semantic segmentation classification, among others. Then the visual encoder is frozen, and an \acrshort{rl} algorithm (Rainbow-IQN Ape-X \cite{Hessel2017RainbowCI}) is used to train the policy network on the latent representation. Ahmed \etal \cite{Ahmed2022} also used the concept of affordances, but went even further and used the affordances themselves as the input of the \acrshort{rl} algorithm. More recently, Zhao \etal \cite{Zhao2022CADREAC} proposed CADRE, a cascade \acrshort{rl} framework for vision-based urban autonomous driving. Their method first trains offline a co-attention perception module to learn relationships between the input images and the corresponding command controls from a driving dataset. This module is then frozen and is used as the input of an efficient distributed \acrfull{ppo} \cite{Schulman2017ProximalPO} that learns the driving policy network online \cite{Zhao2022CADREAC}. The usage of the two-stage training allowed these approaches to use large image encoders to derive a more complex representation of the environment from the sensor data. However, one can argue that the obtained representation may not be totally aligned with the downstream task since it was not trained jointly with the driving policy network. \acrshort{rlfp} aims to fix this limitation by updating the image encoder alongside the driving policy network. It is a method that is receiving massive attention in recent years and can offer numerous benefits to vision-based urban \acrshort{ad}.

\subsection{Reinforcement Learning from Pixels}  

Sample-efficient \acrshort{rl} algorithms capable of training directly from pixel observations could open up a multitude of real-world applications \cite{Kostrikov2020ImageAI}. However, simultaneously training an image encoder and a policy network is a challenging problem due to the strong correlation between samples, sparse reward signal, and degenerated feature representations \cite{Cetin2022StabilizingOD, Kostrikov2020ImageAI, Shelhamer2016LossII}. Naive approaches that use a large image encoder result in severe overfitting, and a smaller image encoder usually produces impoverished representations which limit the performance of the agent \cite{Kostrikov2020ImageAI}. One way of addressing this problem is to employ auxiliary losses. Shelhamer \etal \cite{Shelhamer2016LossII} proposed to use several auxiliary losses to enhance the performance of \acrfull{a3c} \cite{Mnih2016AsynchronousMF}. Zhang \etal \cite{Zhang2020LearningIR} predicted the rewards and dynamics of the environment as auxiliary losses.
Yarats \etal \cite{Yarats2019ImprovingSE} proposed SAC+AE, where the authors demonstrated that combining the off-policy \acrshort{rl} algorithm \acrfull{sac} \cite{Haarnoja2018SoftAO} with pixel reconstruction is vital for learning good representations. Following this line, Srinivas \etal \cite{Srinivas2020CURLCU} proposed CURL -- Contrastive Unsupervised Representations for Reinforcement Learning.  CURL uses contrastive learning to maximize agreement between an augmented version of the same observation, and to minimize agreement between different observations \cite{Srinivas2020CURLCU}. The authors showed that this method significantly improves the sample efficiency of the algorithm. A different line of research was proposed by Kostrikov \etal \cite{Kostrikov2020ImageAI}, where the authors proposed DrQ. This work demonstrated how image augmentations can be applied in the context of model-free off-policy \acrshort{rl} algorithms. The authors proved that using image augmentations leads to better results than using auxiliary losses \cite{Kostrikov2020ImageAI}. Finally, Yarats \etal \cite{Yarats2021MasteringVC} proposed an improved version of DrQ, named DrQ-V2. This version is the result of several algorithmic changes: (i) changing from \acrshort{sac} to \acrfull{ddpg} \cite{Lillicrap2015ContinuousCW}, (ii) incorporating multi-step return, (iii) improving the data augmentation technique, (iv) introducing an exploration schedule, (v) selecting better hyper-parameters \cite{Yarats2021MasteringVC}.

\section{Method}

RLAD is the first \acrshort{rlfp} method applied to the domain of urban \acrshort{ad}. Its main purpose is to derive a feature representation from the sensor data that is fully aligned with the driving task while learning the driving policy simultaneously. The core of RLAD is built upon DrQ \cite{Kostrikov2020ImageAI}, but with several modifications. First, in addition to the image augmentations, we also append at the end of each convolutional layer of the image encoder, a regularization layer called \acrfull{alix} \cite{Cetin2022StabilizingOD} (more details in Section \ref{encoder}), which significantly improves the stability and efficiency of the training. Second, we performed an extensive study of the best hyperparameters, where we realized that some hyperparameters of DrQ weren't optimal for the \acrshort{ad} domain. Finally, we use an additional loss for traffic light classification in order to guide the latent representation of the image ($\tilde{\boldsymbol{i}}$) to contain information about the traffic lights.

\begin{figure*}[t]
\centering
\includegraphics[width=2\columnwidth]{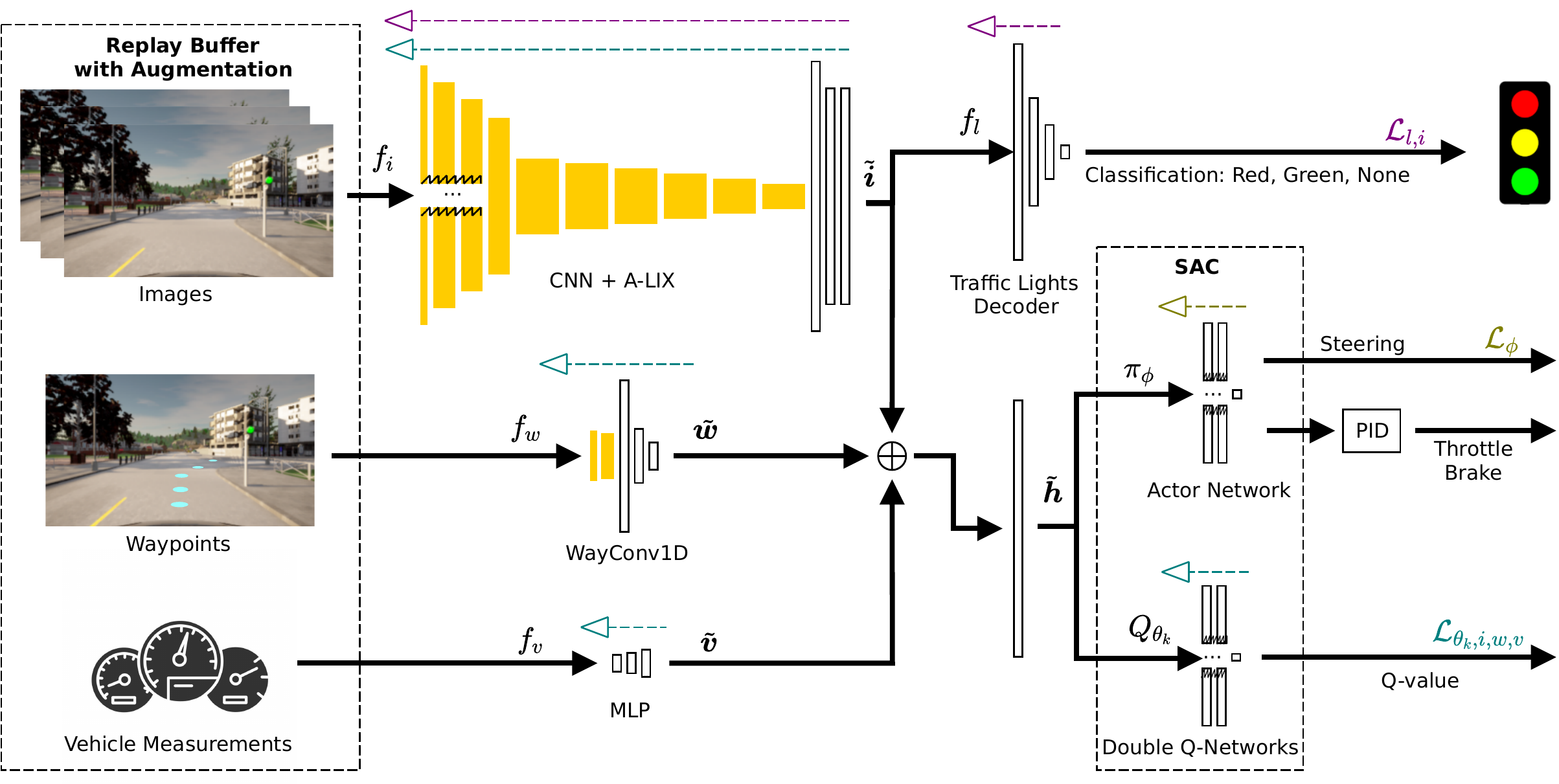}
\caption{Architecture of RLAD. As input, the system receives $K$ consecutive central images, $N$ waypoints computed using a global planner, and measurements from the vehicle from the last $K$ steps. Each input is processed independently by a different encoder. The latent representations of each input are then concatenated, forming the input of the \acrshort{sac} algorithm 
($\tilde{\boldsymbol{h}} = \begin{bmatrix}\tilde{\boldsymbol{i}} & \tilde{\boldsymbol{w}} &\tilde{\boldsymbol{v}}\end{bmatrix}$).
The actor network of the \acrshort{sac}, along with a PID, is responsible for outputting the command controls, while the Q-networks are responsible for outputting the value function. To guide $\tilde{\boldsymbol{i}}$ to contain information about the traffic lights, we add an auxiliary branch to perform traffic light classification. All elements of the neural networks are represented at scale. The dashed arrows provide a visual representation of how each loss function affects the parameters of the system in the backpropagation stage.}
\label{fig:rlad}
\end{figure*}

\subsection{Learning Environment}

The learning environment is defined as a Partially Observable Markov Decision Process (POMDP). The environment was built by using the CARLA driving simulator (version 0.9.10.1) \cite{Dosovitskiy2017CARLAAO}. 

\paragraph{State space} $\mathcal{S}$ is defined by CARLA, containing the ground truth about the world. The agent has no access to the state of the environment.

\paragraph{Observation space} At each step the state $s_t \in \mathcal{S}$ generates the corresponding observation $o_t \in \mathcal{O}$, which is conveyed to the agent. An observation is a stack of $K = 3$ sets of tensors from the last $K$ timesteps. Specifically, $o_t = \{(\pmb{I}, \pmb{W}, \pmb{V})_k\}_{k=0}^2$, where: $\pmb{I}$ is a 3$\times$256$\times$256 image, $\pmb{W}$ corresponds to the 2D coordinates related to the vehicle, for the next $N=10$ waypoints provided by the global planner from CARLA, and $\pmb{V}$ corresponds to a two-dimensional vector containing the current speed and steering of the vehicle. 

\paragraph{Action Space} $\mathcal{A}$ is composed of three continuous actions: throttle, which ranges from 0 to 1; brake, which ranges from 0 to 1; and steering, which ranges from -1 to 1.

\paragraph{Reward function} We used the reward function defined in \cite{Zhang2021EndtoEndUD} because it has been shown to accurately guide the \acrshort{ad} training.

\paragraph{Training} We used CARLA at 10 FPS. Similarly to \cite{Zhang2021EndtoEndUD}, at the beginning of the episode, the start and target locations are randomly generated and the desired route is computed using the global planner. When the target location is reached, a new random target location is computed. The episode is terminated if one of the following conditions is met: collision, running a red traffic light, blocked, or if a predefined timeout is reached.

\subsection{Agent Architecture}

The architecture of RLAD is depicted in Figure \ref{fig:rlad}. In general, our system has three main components: an encoder (Section \ref{encoder}), an \acrshort{rl} algorithm (Section \ref{rl-algorithm}), and an auxiliary loss (Section \ref{auxiliary_loss}). To simplify the longitudinal control and ensure smooth control, we reparameterize the throttle and brake commands using a target speed. As such, a PID controller is appended at the end of the actor network that produces the corresponding throttle and brake commands that match the predicted target speed. 

\subsubsection{Encoder} \label{encoder}

The encoder is responsible for transforming the data from the sensors ($o_t$) into a low-dimension feature vector ($\tilde{\boldsymbol{h}}_t$) to be processed by the \acrshort{rl} algorithm. 

\paragraph{Image Encoder}

As demonstrated in \cite{Kostrikov2020ImageAI}, the size of the image encoder is a critical element in an RLfP method. Due to the weak signal of the \acrshort{rl} loss, encoders commonly used in \acrshort{ad} methods, such as ResNet50 \cite{He2015DeepRL} ($\sim$ 25M parameters) or Inception V3 \cite{Szegedy2015RethinkingTI} ($\sim$ 27M parameters), are impracticable. On the other side, small encoders, designed for scenarios of smaller complexity, such as IMPALA \cite{Espeholt2018IMPALASD} ($\sim$ 0.22M parameters), cannot produce an adequate representation of the environment, which limits the performance of the driving agent. For urban \acrshort{ad}, our findings suggest that the optimal configuration entails a trade-off between larger networks that are unsuitable for training with \acrshort{rl} and smaller networks that cannot accurately perceive the environment. The architecture of the proposed image encoder is shown in Table \ref{tab:encoder}, containing around 1M parameters.
Similarly to DrQ and DrQ-V2, we leverage simple image augmentations to regularize the value function \cite{Kostrikov2020ImageAI, Yarats2021MasteringVC}. First, we apply padding to each side of the 256$\times$256 image by repeating the 8 boundary pixels and then selecting a random crop of 256$\times$256. As in \cite{Yarats2021MasteringVC}, we found it useful to apply bilinear interpolation on the cropped images. In addition to the image augmentations, we also found that appending an A-LIX layer \cite{Cetin2022StabilizingOD} at the end of each convolution layer improves the performance of the agent, possibly by preventing a phenomenon called catastrophic self-overfitting (spatially inconsistent feature maps that lead to discontinuous gradients in the backpropagation). A-LIX is applied on the features produced by the convolution layers $a \in \mathbb{R}^{C\times H \times W}$, by randomly mixing each component $a_{cij}$  with its neighbors belonging to the same feature map.
Consequently, the output of A-LIX is of the same dimensionality as the input, but with the difference that the computation graph minimally disrupts the information of each feature $a_{cij}$, while smoothing discontinuous component of the gradients signal during backpropagation \cite{Cetin2022StabilizingOD}. Hence, this technique works by enforcing the image encoder to produce feature maps that are spatially consistent and thus minimizing the effect of the catastrophic self-overfitting phenomenon. This process can be succinctly summarized as 
$\tilde{\boldsymbol{i}}_t = f_{i}(\mathrm{aug}(\left[\{\pmb{I}_{t-k}\}_{k=0}^{2}\right]))$
, where $f_{i}$ is the image encoder, $\mathrm{aug}$ corresponds to the data augmentation applied, and $\tilde{\boldsymbol{i}}_t$ corresponds to the latent representation of the stack of three consecutive images $\left(\left[\{\pmb{I}_{t-k}\}_{k=0}^{2}\right]\right)$.


\begin{table}[ht]
\centering
\caption{Architecture of the proposed image encoder. After each convolution layer, we applied the ReLU function \cite{Agarap2018DeepLU} and the \mbox{A-LIX} regularizer layer \cite{Cetin2022StabilizingOD}.}
\begin{tabular}{ccc}
\toprule
type       & kernel/stride & input size \\ 
\midrule
$\mathrm{conv}$       & 3$\times$3/2  & 9$\times$256$\times$256 \\ 
$\mathrm{conv}$       & 3$\times$3/2  & 32$\times$127$\times$127   \\ 
$\mathrm{conv}$       & 3$\times$3/2  & 32$\times$63$\times$63   \\ 
$\mathrm{conv}$       & 3$\times$3/2  & 32$\times$31$\times$31   \\ 
$\mathrm{conv}$       & 3$\times$3/1  & 64$\times$15$\times$15  \\ 
$\mathrm{conv}$       & 3$\times$3/1  & 64$\times$13$\times$13   \\ 
$\mathrm{conv}$       & 3$\times$3/1  & 64$\times$11$\times$11     \\ 
$\mathrm{conv}$       & 3$\times$3/1  & 64$\times$9$\times$9     \\ 
$\mathrm{conv}$       & 3$\times$3/1  & 64$\times$7$\times$7     \\ 
$\mathrm{flatten}$    & -             & 64$\times$5$\times$5      \\ 
$\mathrm{linear}$     & -             & 1600        \\ 
$\mathrm{layer norm}$ & -             & 256        \\ 
$\mathrm{tanh}$       & -             & 256        \\ 
\bottomrule
\end{tabular}
\label{tab:encoder}
\end{table}

\paragraph{Waypoint Encoder}

Usually, the waypoint encoder consists of using the mean orientation between the current pose of the agent and the next $N$ waypoints \cite{Agarwal2021} or flattening the waypoints' 2D coordinates into a vector and then applying an MLP \cite{Cai2020ProbabilisticEV}. In our point of view, both approaches have serious limitations. The former approach clearly oversimplifies the problem by encoding all waypoint coordinates into a single value. This method only works for small values of $N$, because as $N$ increases, the waypoints become more scattered, and thus the  mean orientation ceases to be a reliable indicator. Although the latter approach works for all values of $N$, by flattening the 2D waypoint coordinates into a vector, the 2D geometrical information is not being used. To overcome both limitations, we propose WayConv1D, a waypoint encoder that leverages the 2D geometrical structure of the input by applying 1D convolutions with a 2$\times$2 kernel over the 2D coordinates of the next $N$ waypoints. The output of the 1D convolution is then flattened and processed by an MLP. This process can be summarized as $\tilde{\boldsymbol{w}}_t = f_{w}(\pmb{W}_t)$, where $f_{w}$ corresponds to the WayConv1D, and $\tilde{\boldsymbol{w}}_t$ corresponds to the latent representation of the waypoints for the current step ($\pmb{W}_t$).  We found that with WayConv1D, the agent learns more efficiently to follow the trajectory without oscillating near the center of the lane. This is a common issue encountered when utilizing \acrshort{rl} in the urban \acrshort{ad} domain, as documented in previous studies \cite{Zhang2021EndtoEndUD, Toromanoff2019EndtoEndMR}.

\paragraph{Vehicle Measurements Encoder}

Similarly to \cite{Cai2020ProbabilisticEV}, we apply an MLP to the vehicle measurements: $\tilde{\boldsymbol{v}}_t = f_{v}(\left[\{\pmb{V}_{t-k}\}_{k=0}^{2}\right])$, where $f_v$ is the MLP, and $\tilde{\boldsymbol{v}}_t$ corresponds to the latent representation of the concatenation of the vehicle measurements across three steps $\left(\left[\{\pmb{V}_{t-k}\}_{k=0}^{2}\right]\right)$.

\subsubsection{\acrshort{rl} Algorithm} \label{rl-algorithm}

As the \acrshort{rl} algorithm, we use the \acrshort{sac} \cite{Haarnoja2018SoftAO}, which is a model-free off-policy actor-critic algorithm that learns two Q-functions $Q_{\theta_1}$, $Q_{\theta_2}$, a stochastic policy $\pi_\phi$, and a temperature $\alpha$ to find an optimal policy by optimizing a $\gamma$-discounted maximum-entropy objective \cite{Ziebart2008MaximumEI, Kostrikov2020ImageAI}. The actor policy $\pi_{\phi}(a_t \mid \tilde{\boldsymbol{h}}_t)$ is a parametric $\mathrm{tanh}$-Gaussian that given $\tilde{\boldsymbol{h}}_t = \begin{bmatrix}\tilde{\boldsymbol{i}}_t & \tilde{\boldsymbol{w}}_t &\tilde{\boldsymbol{v}}_t\end{bmatrix}$,  samples $a_t = \mathrm{tanh}\left(\mu_{\phi}(\tilde{\boldsymbol{h}}_t) + \sigma_{\theta}(\tilde{\boldsymbol{h}}_t)\epsilon\right)$, where $\epsilon \sim \mathcal{N}(0,1)$, and $\mu_{\phi}$ and $\sigma_\phi$ are the parametric mean and standard deviation. The double Q-networks are learned by optimizing a single step of the soft Bellman residual:

\begin{equation}\label{eq:critic}
\begin{split}
    \mathcal{L}_{\theta_k, i, w, v} = \mathbb{E}_{\substack{o_t,a_t,o_{t+1} \sim \mathcal{D} \\ a_{t+1}^{\prime} \sim \pi_\phi(\cdot \mid \tilde{\boldsymbol{h}}_{t+1})}} \left[ \left( Q_{\theta_k}\left( \tilde{\boldsymbol{h}}_t, a_t \right) - y \right)^2  \right], \\ \forall k \in \{1,2\},
\end{split}
\end{equation}
with the TD target $y$ defined as:
\begin{equation}
\begin{split}
    y = r_t + \gamma \biggl(\smash{\displaystyle\min_{k=1,2}}Q_{\bar{\theta}_k} \left( \tilde{\boldsymbol{h}}_{t+1}, a_{t+1}^{\prime} \right) - \\ \alpha \log \pi_\phi \left(a_{t+1}^{\prime} \mid  \tilde{\boldsymbol{h}}_{t+1}\right) \biggr),
\end{split}
\end{equation}
where $\mathcal{D}$ represents the replay buffer, $r_t$ is the reward received, $\gamma$ is the discount factor, and $Q_{\bar{\theta}_1}$ and $Q_{\bar{\theta}_2}$ denote the \acrfull{ema} of the parameters of $Q_{\theta_1}$ and $Q_{\theta_2}$, respectively. Similarly to DrQ-V2, we found it useful to use a single encoder, rather than a main encoder and an \acrshort{ema} of the main encoder. The policy is updated to maximize the expected future return plus the expected future entropy:
\begin{equation}\label{eq:actor}
\begin{split}
    \mathcal{L}_{\phi} = -\mathbb{E}_{\substack{{o_t \sim \mathcal{D}} \\ a_t \sim \pi_\phi(\cdot \mid \tilde{\boldsymbol{h}}_t)}} \biggl[ \smash{\displaystyle\min_{k=1,2}} Q_{\theta_k} \left( \tilde{\boldsymbol{h}}_t, a_t\right) - \\ \alpha \log \pi \left(a_t \mid \tilde{\boldsymbol{h}}_t\right) \biggr].
\end{split}
\end{equation}
Finally, the temperature $\alpha$ is learned using the loss proposed by Haarnoja \etal \cite{Haarnoja2018SoftAA}.

As noted in Equation \ref{eq:critic} and \ref{eq:actor}, not all losses are propagated to the encoders. Following \cite{Yarats2019ImprovingSE}, we block the actor's gradients
from propagating to the encoder. 
In contrast with DrQ-V2, we found that using a learning rate of $10^{-3}$, instead of $10^{-4}$, results in a faster and more stable training. One intuition to explain this improvement is related to the observation space. DrQ-V2 was evaluated in tasks where the observation space is likely task-relevant, whereas in \acrshort{ad} the observation space contains task-irrelevant information, such as clouds and buildings. Thus, with a larger learning rate, the encoder will learn faster to distinguish the task-relevant objects from the non-relevant, which prevents the agent from exploring using unreliable representations of the environment.

\subsubsection{Auxiliary Loss} \label{auxiliary_loss}

Based on initial experiments, we observed that the agent struggled to associate the color of the traffic light with the negative reward incurred when passing through a red light. This is understandable, particularly when we take into account that the traffic light color occupies only a small fraction of the entire image. To address this issue, we implemented an auxiliary loss that strengthens the significance of traffic light information in the latent representation of the image ($\tilde{\boldsymbol{i}}$). As such, we added a traffic light decoder ($f_{l}$) to the end of the image encoder and perform traffic light classification using three classes ($C=3$): \textit{None}, \textit{Red}, and \textit{Green}. \textit{None} signifies that there is no traffic light within the vicinity of the agent, \textit{Red} indicates the presence of a red or yellow traffic light near the agent, and finally, \textit{Green} denotes a green traffic light near the agent. Every time we sample a batch of transitions from the replay buffer, we perform traffic light classification using the cross-entropy loss:
\begin{equation}
    \mathcal{L}_{l,i} = - \sum_{b=1}^{B}\sum_{c=1}^{C} \log\left(\frac{e^{\left(x_{b,c}\right)}}{\sum_{i=1}^{C}e^{\left(x_{b,i}\right)}}\right) y_{b,c},
\end{equation}
where $B$ is the batch size, $x$ corresponds to the logits outputted by $f_l$, and $y$ corresponds to the ground truth class. In the backpropagation stage, this loss updates both the parameters of the traffic light decoder and the parameters of the image encoder.

\section{Experiments}

In this section, we compare RLAD with the state-of-the-art RLfP methods, applied to the domain of urban AD. First, we define the setup of the experiments, and then compare RLAD with the state-of-the-art methods both in terms of expected return and using specific metrics related to urban AD. Finally, we present an ablation study that guided the development of RLAD.

\subsection{Setup}

\paragraph{Benchmark} The methods are evaluated on the NoCrash benchmark \cite{Codevilla2019ExploringTL}. This benchmark considers generalization from Town 1, a town composed of one-lane roads and T-junctions with traffic lights, to Town 2, which is a scaled-down version of Town 1 with different textures. The training is performed using four training weather types, and the testing uses two different weather types. This benchmark has three levels of traffic density (empty, regular, and dense) according to the number of vehicles and pedestrians. The results are reported in terms of success rate, which is the percentage of routes completed without collision. Additionally, we also report information related to the percentage of route completion, red light infractions, collisions with vehicles, pedestrians, and layout, and blockages per kilometer.

\paragraph{Training Details} All algorithms are trained on the same hardware: a single NVIDIA RTX 2080 Ti. The algorithms are trained for $10^6$ environment steps and are evaluated every \num{20000} environment steps. Each evaluation query averages episode returns over 10 episodes. The Deep Learning library used was PyTorch \cite{Paszke2019PyTorchAI}. The list with the main hyperparameters used is present in Table \ref{tab:hyperparameters}.

\begin{table}[ht]
\centering
\caption{List of the hyperparameters used by RLAD.}
\begin{tabular}{lc}
\toprule
Parameter                                                  & Value                                       \\ 
\midrule
Replay Buffer capacity                                     & 100000                                      \\ 
Batch size                                                 & 256                                         \\ 
Action repeat                                & 2 
                           \\

Discount factor $\gamma$                                       & 0.99                                        \\
Optimizer                                                  & Adam  \cite{Kingma2014AdamAM}                                      \\ 
Learning rate                                              & $10^{-3}$                                        \\
Critic target update frequency                             & 1                                           \\ 
Critic Q-function soft update rate                         & 0.01                                        \\ 
Critic update frequency                                    & 1 \\
Actor update frequency                                     & 1                                           \\ 
Auxiliary loss update frequency                            & 1 \\
$\mathrm{dim}(\tilde{\boldsymbol{i}})$                                                     & 256                                         \\ 
$\mathrm{dim}(\tilde{\boldsymbol{w}})$                                                     & 32                                          \\ 
$\mathrm{dim}(\tilde{\boldsymbol{v}})$                                                     & 16                                           \\ 
\acrshort{sac} networks size                                           & 1024                                        \\ 
Actor log stddev bounds                                                          & [-10,2]                                      \\ 
Init temperature                                                 & 0.1                                         \\ 

\bottomrule \\
\end{tabular}
\label{tab:hyperparameters}
\end{table}

\paragraph{Baselines}  Given that we are the first to propose an RLfP method applied in urban AD, we compare our method with the state-of-the-art RLfP methods: SAC+AE \cite{Yarats2019ImprovingSE}, CURL \cite{Srinivas2020CURLCU}, DrQ \cite{Kostrikov2020ImageAI}, and DrQ-V2 \cite{Yarats2021MasteringVC}. The official implementation of these methods only uses images as input, so we added two additional encoders: a waypoint encoder similar to \cite{Cai2020ProbabilisticEV}, and a vehicle measurement encoder similar to ours. These encoders were selected because they are the most commonly utilized in the end-to-end \acrshort{ad} field. For a fair comparison, we also reparameterize the throttle and brake commands using a PID controller.

\subsection{Comparison with Baselines}

\begin{figure}[t]
\centering
\includegraphics[width=1.0\columnwidth]{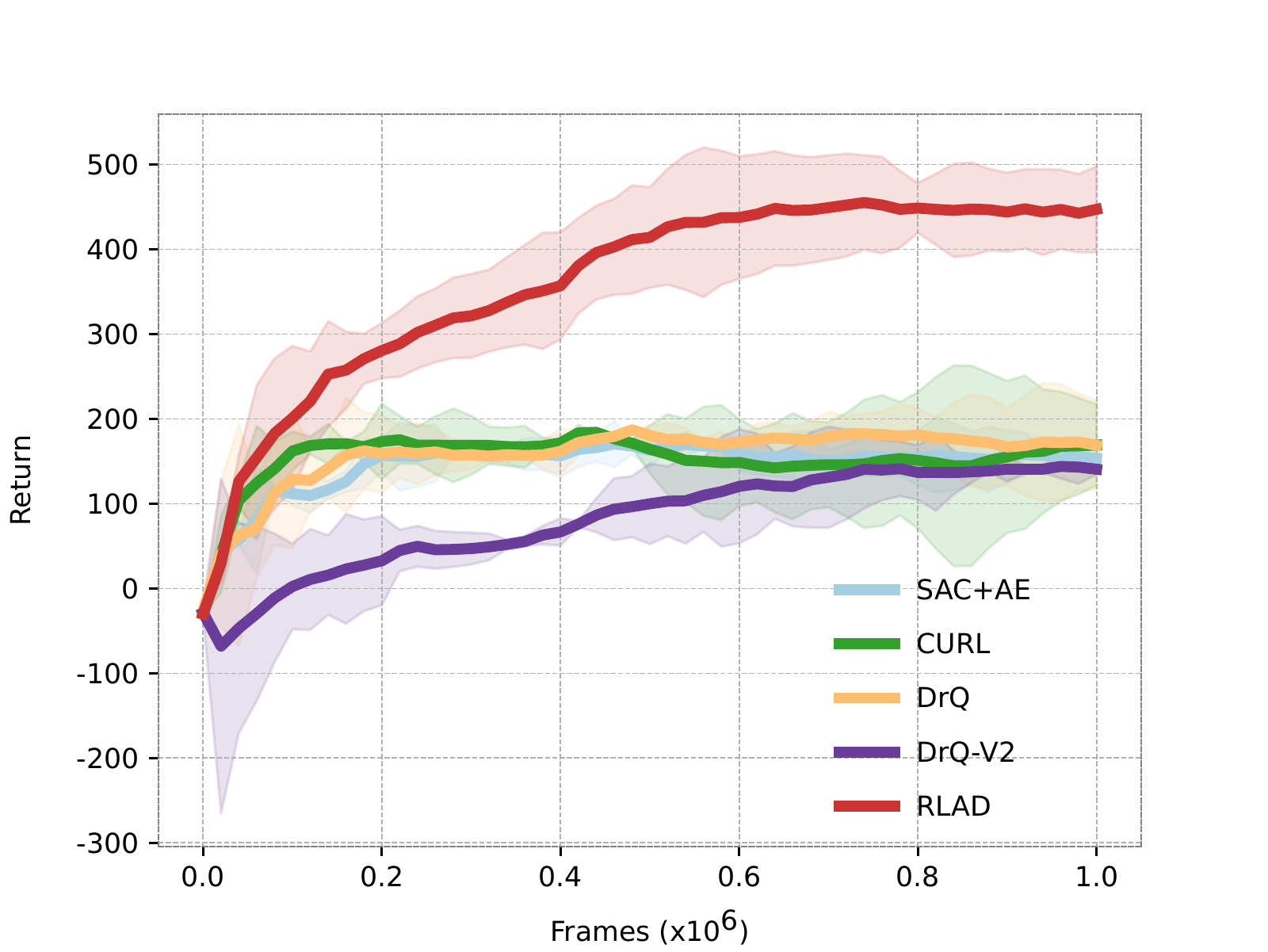}
\caption{Comparison of RLAD with state-of-the-art RLfP methods in terms of average return per episode on the NoCrash benchmark. The solid lines represent the mean performance over 3 seeds, and the shaded regions represent \SI{90}{\percent} confidence intervals.}
\label{fig:reward}
\end{figure}

Figure \ref{fig:reward} depicts the average return for each method during the NoCrash benchmark's training process. It is evident that RLAD significantly outperforms all state-of-the-art methods. By the end of the training, RLAD manages to attain an average return that is roughly three times greater than all other methods.

Table \ref{tab:nocrash_testing} shows the performance of the algorithms in terms of success rate for every task of the NoCrash benchmark under testing conditions. In the empty task, all algorithms perform reasonably well, with the exception of DrQ-V2. However, as the difficulty of the task increases, the difference between the performance of RLAD and the other methods becomes more evident. Although RLAD performs equally to DrQ in the empty task, it outperforms the second best method by \SI{50}{\percent} in the regular task and by \SI{220}{\percent} in the dense task.

\begin{table}[ht]
\centering
\caption{Success rate (\si{\percent}) on NoCrash benchmark for each task in testing conditions (Town 2 with new weather). The results for each method correspond to the best seed considering the average episode return (Figure \ref{fig:reward}). Best scores for each task highlighted in bold.}
\begin{tabular}{lccc}
\toprule                                     
                                 & Empty & Regular & Dense \\ \midrule
SAC+AE                       &   82    &    42     &   6    \\ 
CURL                          &    74   &    30     &   2    \\ 
DrQ                           &    \textbf{94}   &    42     &   10    \\
DrQ-V2                         &    10   &     8    &    0   \\ 
RLAD                           &    \textbf{94}   &     \textbf{62}    &    \textbf{32}   \\ \bottomrule \\
\end{tabular}
\label{tab:nocrash_testing}
\end{table}

Table \ref{tab:infraction} provides a detailed analysis of the performance of the methods with respect to \acrshort{ad}-related metrics, specifically through an infraction analysis conducted on the regular task of the NoCrash benchmark. With the exception of DrQ-V2, all state-of-the-art \acrshort{rlfp} methods achieve a route completion over \SI{90}{\percent}, but achieve a success rate of less than \SI{50}{\percent}. This clearly shows that the state-of-the-art \acrshort{rlfp} methods are very good at following the trajectory generated by the global planner, but struggle to deal with dynamic obstacles, as is the case of vehicles and pedestrians. In contrast, RLAD is capable of dealing with dynamic obstacles, achieving the best score for all metrics related to collisions. The scores related to the red light infractions clearly demonstrate that obeying the traffic lights is a challenging task for \acrshort{rlfp} algorithms. RLAD with the auxiliary loss is able to perform \SI{17}{\percent} better when compared with the second best, but still very poorly when compared with state-of-the-art methods that use \acrshort{il} \cite{Chitta2022TransFuserIW, Shao2022SafetyEnhancedAD, Chen2022LearningFA}. This limitation arises from the impracticability of using large image encoders in \acrshort{rlfp}, which makes it challenging to create representations of the environment that include small yet important features, such as the color of traffic lights. Among all methods, DrQ-V2 is the one that performs worse in virtually all metrics. Internal investigations showed that the primary reason for this performance was the \acrshort{rl} algorithm used - \acrshort{ddpg}. Using our training conditions, \acrshort{ddpg} quickly converges to a suboptimal policy, where the agent tends to remain still in various situations. This problem can be easily identified in the column related to the blockages of DrQ-V2: 22.14 blockages per kilometer.

Although RLAD outperforms all \acrshort{rlfp} methods in the urban \acrshort{ad} domain, it is not yet competitive with state-of-the-art \acrshort{rl} methods that decouple the training of encoder and the policy network \cite{Toromanoff2019EndtoEndMR, Zhao2022CADREAC, Ahmed2022}. However, in the field of continuous control tasks in the MuJoCo simulator \cite{Tassa2018DeepMindCS}, \acrshort{rlfp} methods have already surpassed those that decouple the encoder from the policy network, which suggests that the same pattern may occur in the urban \acrshort{ad} domain as well.
Furthermore, the image encoder of state-of-the-art \acrshort{rl} methods that decouple the encoder from the policy network contains around 25 times more parameters than the image encoder of RLAD \cite{Toromanoff2019EndtoEndMR, Zhao2022CADREAC, Ahmed2022}, requiring more computation power and resources, which may compromise their application in real-time settings.

\begin{table*}[ht]
\centering
\caption{Driving performance and infraction analysis on the NoCrash benchmark, using the regular task in testing conditions. The results for each method correspond to the best seed considering the average episode return (Figure \ref{fig:reward}). Best scores highlighted in bold.}
\begin{tabular}{lccccccc}
\toprule

        & \makecell[c]{Success \\ rate} & \makecell[c]{Route \\ completion} & \makecell[c]{Collision \\ pedestrian} & \makecell[c]{Collision \\ vehicle} & \makecell[c]{Collision \\ layout} & \makecell[c]{Red light \\ infraction}  & \makecell[c]{Agent \\ blocked}\\ 
        & \makecell[c]{\si{\percent}, $\uparrow$}           & \makecell[c]{\si{\percent},$\uparrow$}               & \makecell[c]{\#/Km, $\downarrow$}           & \makecell[c]{\#/Km, $\downarrow$}                & \makecell[c]{\#/Km, $\downarrow$}  & \makecell[c]{\#/Km $\downarrow$}             & \makecell[c]{\#/Km, $\downarrow$}                \\ \midrule
SAC+AE  &       42   &           98       &        0.60      &        1.71            &     0.99           &       6.16        &   0.23      \\ 
CURL    &     30       &       99           &     0.97        &       1.91        &       1.68           &      6.81       &   0.20    \\  
                                                         
DrQ     &      42       &        \textbf{100}         &        0.77         &      1.51                &       0.19           &     7.35        &     \textbf{0.00}    \\   DrQ-V2            &      8       &          53       &        0.58         &         1.68             &       1.63            &    7.04       &     22.14    \\ 
RLAD    &    \textbf{62}      &      94          &      \textbf{0.41}           &           \textbf{0.71}          &     \textbf{0.16}       &    \textbf{5.10}         &     0.84  \\                                             
\bottomrule \\
\end{tabular}
\caption{Driving performance and infraction analysis on the NoCrash benchmark, using the regular task in testing conditions. The results for each method correspond to the best seed considering the average episode return (Figure \ref{fig:reward}). Best scores highlighted in bold.}
\label{tab:infraction}
\end{table*}

\subsection{Ablation Study}

\begin{figure}[t]
\centering
\includegraphics[width=1.0\columnwidth]{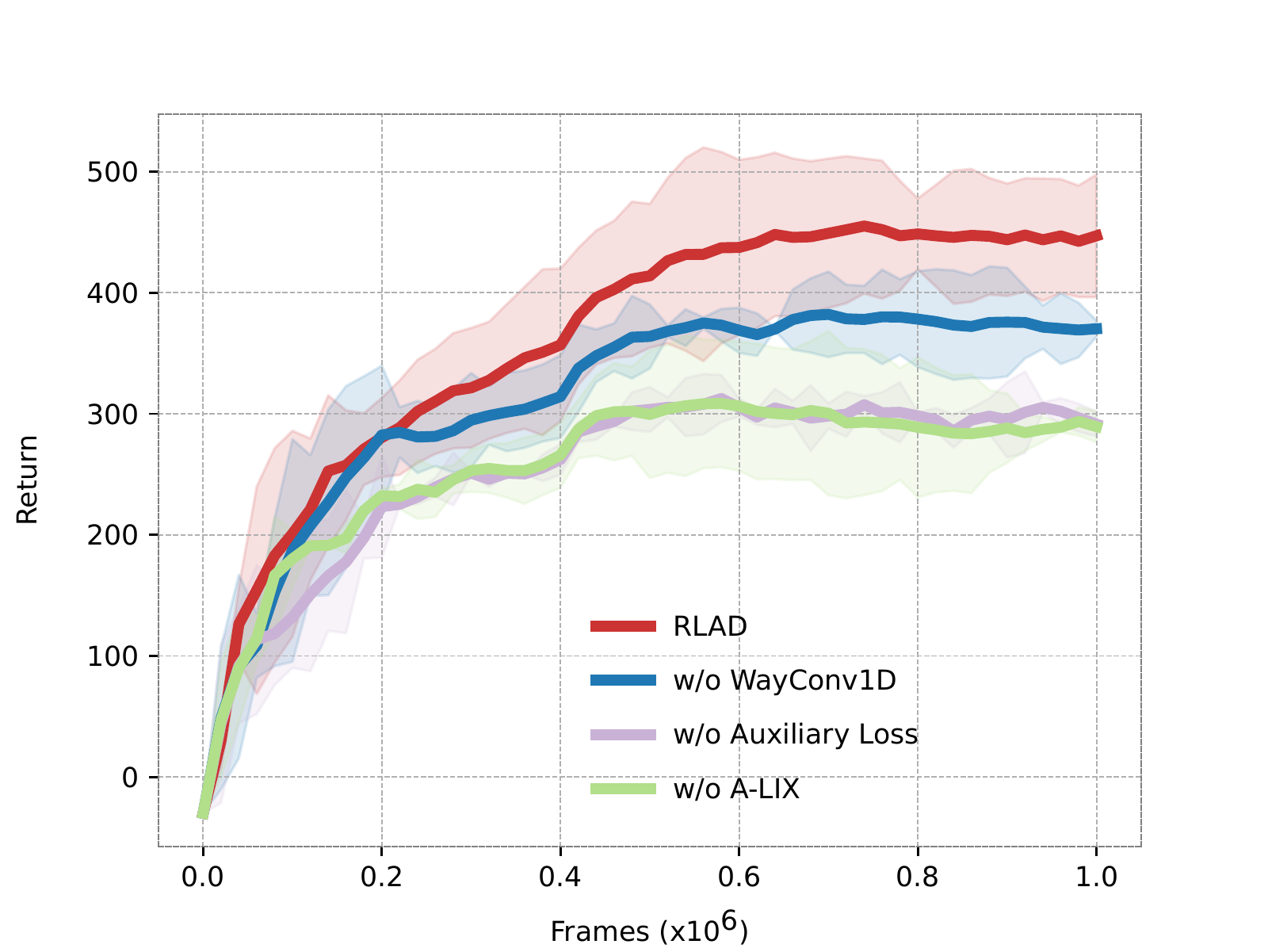}
\caption{Ablation study that guided the development of RLAD. The solid lines represent the mean performance over 3 seeds, and the shaded regions represent \SI{90}{\percent} confidence intervals.}
\label{fig:ablation}
\end{figure}

We performed three experiments: using the waypoint encoder of \cite{Cai2020ProbabilisticEV} instead of WayConv1D; removal of the auxiliary loss; and removal of the A-LIX layers. Figure \ref{fig:ablation} shows the influence of each of these components in terms of the average return per episode. The most impactful components are the auxiliary loss and the A-LIX layers. Removing them results in a performance decrease of \SI{33}{\percent}. Replacing the WayConv1D by the waypoint encoder of \cite{Cai2020ProbabilisticEV} results in a performance decrease of \SI{15}{\percent}. Furthermore, this replacement also leads to more oscillations near the center lane, resulting in a less comfortable driving experience.


\section{Conclusion}

This paper introduced RLAD, the first algorithm that learns simultaneously the encoder and the driving policy network using \acrfull{rl} in the domain of vision-based urban \acrfull{ad}. Our method significantly outperforms all RLfP state-of-the-art methods in this domain. Although our method is not yet competitive with the state-of-the-art methods in end-to-end urban AD, we believe that RLAD can foster the interest in applying RLfP to the domain of urban \acrshort{ad}. Methods that learn simultaneously the encoder and the policy network have demonstrated better performance in the continuous control tasks in the MuJoCo simulator, compared to those that decouple the encoder from the policy network. Based on this information, we have grounds to expect that a comparable pattern will emerge in the realm of urban AD, and we believe that RLAD constitutes the first step toward this end.

\section*{Acknowledgements}
This work has been supported by FCT - Foundation for Science and
Technology, in the context of Ph.D. scholarship 2022.10977.BD and by National Funds through the FCT - Foundation for Science and Technology, in the context of the project UIDB/00127/2020.

{\small
\bibliographystyle{IEEEtran}
\bibliography{egbib}
}

\newpage

\vfill

\end{document}